\pdfoutput=1

\documentclass[11pt]{article}

\usepackage[preprint]{acl}

\usepackage{times}
\usepackage{latexsym}

\usepackage[T1]{fontenc}

\usepackage[utf8]{inputenc}

\usepackage{microtype}

\usepackage{inconsolata}

\usepackage{graphicx}

\usepackage{booktabs} 
\usepackage{multirow}
 \usepackage{array,multirow,graphicx}
 \usepackage{float}

\newcommand{\citeplanguageresource}[1]{\citep{#1}}

\newcommand{\githubUrl}{\url{https://github.com/malteos/turkish-lm-bias}}

\title{Investigating Gender Bias in Turkish Language Models}

\author{Orhun Caglidil, Malte Ostendorff, Georg Rehm \\
German Research Center for Artificial Intelligence (DFKI)\\
         Berlin, Germany \\
         \texttt{first.lastname@dfki.de} \\}

\begin{document}
\maketitle
\begin{abstract}
Language models are trained mostly on Web data, which often contains social stereotypes and biases that the models can inherit. This has potentially negative consequences, as models can amplify these biases in downstream tasks or applications. However, prior research has primarily focused on the English language, especially in the context of gender bias. In particular, grammatically gender-neutral languages such as Turkish are underexplored despite representing different linguistic properties to language models with possibly different effects on biases. In this paper, we fill this research gap and investigate the significance of gender bias in Turkish language models. We build upon existing bias evaluation frameworks and extend them to the Turkish language by translating existing English tests and creating new ones designed to measure gender bias in the context of Türkiye. Specifically, we also evaluate Turkish language models for their embedded ethnic bias toward Kurdish people. Based on the experimental results, we attribute possible biases to different model characteristics such as the model size, their multilingualism, and the training corpora. We make the Turkish gender bias dataset publicly available.
\end{abstract}

\section{Introduction}
\label{sect:intro}

While the growing size of pre-trained language models, such as BERT \citep{devlin2018bert}, has led to large improvements in a variety of natural language processing (NLP) tasks, the success of these models comes with a price: 
They are trained on drastic amounts of mostly Web-based data, which are often prone to social stereotypes and biases that the models might inherit in their training. 
This can have negative consequences, as models can reproduce these biases in downstream tasks or applications \citep{tal2022fewer}. 
For instance, some models predict higher emotion intensity for sentences with female words than male words under the same context \citep{parasurama2021degendering}. 

Another application exemplifying the embedded cultural stereotypes is statistical machine translation, as shown by \citep{caliskan2017semantics}. 
Translations to English from a gender-neutral language such as Turkish, which does not have any grammatical gender like the gendered pronouns `\textit{he}' or `\textit{she}' in English, lead to gender-stereotyped sentences. 
For instance, Google Translate converts these Turkish sentences with gender-neutral pronouns: 
`\textit{O bir doktor. O bir hemşire.}' to the
English sentences: `\textit{He is a doctor. She is a nurse.}' 
The same behavior can be observed when translating the above two Turkish sentences into other commonly spoken languages with grammatical gender like Spanish, Russian, German, and French. 
The gender-neutral Turkish pronoun `\textit{o}' is converted into gender-stereotyped pronouns in the respective language in every case.


Bias could be discussed from a descriptive or a normative perspective. 
Statistics dividing occupations by gender show that 19.3\% of executive positions in Turkey
in 2020 were held by women.\footnote{\url{https://data.tuik.gov.tr/Bulten/Index?p=Istatistiklerle-Kadin-2021-45635} (TÜİK Kurumsal)}  
Respectively, the statistical probability that an executive is female is lower than for a man
and if a language model estimates p(`woman' | `executive') lower than p(`man'
| `executive') it is descriptively correct. 
However, these gender gaps in executive positions have historically resulted from societal gender inequalities and culturally established gender roles, as well as women's limited access to education \citep{ozaydinlik2014toplumsal}. 
The perpetuation of this stereotype in language helps to maintain gender inequality \citep{blodgett2020language}. 
Therefore, most works on equity in NLP take a normative stance. 
The premise is that algorithms should not model stereotypes that threaten to perpetuate social inequalities \citep{kraft2021triggering}. 
Mitigating different types of bias in LMs would have diverse implications: 
It would allow us to avoid amplifying these biases. 
But also, by avoiding algorithms enforcing social biases
against minorities, one could shift the social balance in the long term.

Despite attracting much attention, it remains still unclear what the full capabilities but also limitations of LMs are. Previous research in this regard has primarily focused on the English language, especially in the content of gender bias in language models.
However, the investigation of more languages with different linguistic elements than English, especially the ones like Turkish that are grammatically gender-neutral, can deepen our insights into the role of gender bias in LMs.
This research gap shall be addressed in this paper. 
In our work, we attempt to investigate the significance of gender bias in Turkish language models.

%
%
The purpose of this research is to use existing evaluation frameworks on Turkish datasets for measuring gender-bias and then use the outcome to evaluate some of the publicly available language models. A comprehensive quantitative evaluation of the most common Turkish LMs, both monolingual and multilingual ones, including multilingual BERT \citep{devlin2018bert}, BERTurk \citeplanguageresource{stefan_schweter_2020_3770924}, mT5 \citep{xue2020mt5} will be conducted.

The quantitative evaluation of the LMs will be complemented with a detailed qualitative evaluation. 
The qualitative evaluation should deepen the gained insights in a qualitative manner. 
Its objective is to answer the two research questions:

\paragraph{(1) What kind of gender stereotypes do the language models pick up? What are some common patterns for the errors that LMs make in this regard?}
The qualitative analysis will attempt to detect the gender stereotypes that models have picked up to identify the similarities and differences when it comes to the success of the downstream performance of the models.

\paragraph{(2) How can these patterns be explained by model properties?} 
Secondly, the research will try to identify the possible roots of these patterns. More specifically, the model properties and architectures will be investigated further with the incentive to find a correlation and causality with the results from the above-mentioned questions.

%

  


\section{Related Work} 


From a quantitative perspective, the systematic differences between a sample and a
population \citep{kraft2021triggering} define a bias. 
The underrepresentation of a specific social group in the training data can affect model performance systematically
\citep{buolamwini2018gender} or cause misrepresentations \citep{blodgett2020language}.
Since LMs are trained on many text corpora that exhibit socially problematic
biases, the models have been shown to capture stereotypical biases \citep{nadeem2020stereoset, may2019measuring, caliskan2017semantics}.
Associations between certain social groups and certain traits are maintained regardless of their databases such that historically created stereotypes persist in society. 
Language plays a crucial role in encoding and transmitting stereotypes and thus creating a kind of consensus within and about certain groups \citep{beukeboom2019stereotypes}.

When natural language becomes the training data, the risk of creating a biased representation of the world arises because the encoded relationship between groups and attributes is misrepresentative. This way, biases can enter statistical models \citet{blodgett2020language}. 
In addition, this phenomenon is particularly likely if the language in the training predominantly reflects the shared perceptions of one social group: For instance, GPT-2 was essentially trained on data from Reddit \citep{radford2019language}, a platform that is mostly dominated by white male users between ages 18 and 29. 
This data, thus, contains significant amounts of white supremacist and misogynistic content \citep{bender2021dangers}.

\citet{brown2020language} have shown in their investigation of gender bias in GPT-3 the associations between gender and occupation. 
They found
that occupations generally have a higher probability of being followed by a male gender identifier than a female one.
\citeauthor{brown2020language} also performed co-occurrence tests, analyzing which words are likely to occur in the vicinity of other preselected words. 
They analyzed the most favored descriptive words for the model along with the number of coincidences each
word co-occurred with a pronoun indicator. 
`Most Favored' here indicates the words that are most aligned with a category by co-occurring with it at a higher rate as compared to the other category. 
The top 10 most biased male descriptive words include `large', `personable', and `lazy' whereas the most biased female descriptive words include `beautiful', `gorgeous', `petite', and `bubbly'.
For a more comprehensive overview of bias in language models and mitigation approaches, we refer to the surveys from \citet{sun-etal-2019-mitigating} and \citet{gallegos2023bias}.



\section{Methodology}
To evaluate an LM for its gender bias, this paper follows the methodology of \citet{may2019measuring}.

\subsection{Evaluation Framework}

\paragraph{Word Embeddings Association Test.}
 
The Word Embeddings Association Test (WEAT), as proposed by \citet{caliskan2017semantics}, is a statistical measure for the association strength between
a pair of word vectors. 
The WEAT was designed after the Implicit Association Test (IAT) \citep{greenwald1998measuring}. IAT is a psychological test that measures human biases by comparing participants' reaction times when pairing concepts that they perceive as similar or as dissimilar. \citet{caliskan2017semantics} use the distance between a pair of vectors of word embeddings which are the semantic representation of words in LMS. The distance is measured by their cosine similarity score, a measure of correlation, as analogous to reaction time in the IAT. \citet{may2019measuring} have demonstrated that female names -- as opposed to male names -- are more strongly associated with family-related attributes in comparison to career-related ones. 
Furthermore, African-American names are more strongly associated
with attributes representing unpleasantness than European-American names.

Both IAT and WEAT use two lists of target words and two lists of attribute
words, the first pair of lists correspond to terms to be compared and the second pair of lists represent
the categories where the presence of bias is anticipated.
Let $X$ and $Y$ be equal-size sets of target
words (e.g., names such as Emily / Keisha) and let 
$A$ and $B$ be sets of attribute
words (e.g., pleasant words like love or peace / unpleasant words like evil or murder). 
The test statistic is a difference
between sums over the respective target words,

\begin{equation}
    \centering
    s(X, Y, A, B) = \sum_{\textit{x}\in X} s(\textit{x}, A, B) -  \sum_{\textit{y}\in Y} s(\textit{y}, A, B)
\end{equation}
where each addend is the difference between mean
cosine similarities of the respective attributes,

 \begin{equation}
    \centering
    s(\textit{w}, A, B) = \frac{\sum_{\textit{A}\in A}cos(\textit{w}, \textit{a})}{|A|} - \frac{\sum_{\textit{b}\in B}cos(\textit{w}, \textit{b})}{|B|}
\end{equation}\\

Additionally, a permutation test on s($X$, $Y$, $A$, $B$) is used to
compute the significance of the association between (A, B) and (X, $Y$ ),
 \begin{equation}
    \centering
    p = Pr[s(X_i, Y_i, A, B) > s(X, Y, A, B)]
\end{equation}
where the probability is computed over the space
of partitions (X$_i$, Y$_i$) of $X$ $\cup$ $Y$ such that X$_i$ and Y$_i$ are of equal size, and a normalized difference of
means of s(\textit{w}, $A$, $B$) is used to measure the effect size -- in other words the magnitude of the association;   

\begin{equation}
    \centering
    d = \frac{\frac{\sum_{\textit{a}\in A} cos(\textit{w}, \textit{a})}{|A|} -  \frac{\sum_{\textit{b}\in B} cos(\textit{w}, \textit{b})}{|B|} }{\sigma_{\textit{w}\in X \cup Y}(\textit{w}, A, B)}
\end{equation} \\

Controlling for significance, a larger effect size indicates a stronger level of bias \citep{may2019measuring, caliskan2017semantics}.

As defined by \citeauthor{caliskan2017semantics}, ten tests using WEAT to measure the bias in different categories have been proposed. 
In this paper, the WEAT lists of words used in the tests were translated into Turkish and modified
accordingly, which means the words in these lists remain associated only with the corresponding category.

\paragraph{Sentence Encoder Association Test.}
Inspired by WEAT, \citet{may2019measuring} introduced The Sentence Encoder Association Test (SEAT). 
SEAT compares sets of sentences, rather than sets of words, by applying WEAT to the vector representation of a sentence. 
This method can be used in contextualized word embeddings like BERT, which is a word embeddings technique that takes into consideration the context of the word and build a vector for each word conditioned on its context.
First, every word in WEAT is replaced by multiple
sentences using a set of semantically bleached sentence templates. Then the same formulas are used as in WEAT where the embeddings represent the entire sentence instead of only a word. 
This approach hypothesizes that the models that use context to obtain more accurate vector representations should not be tested on a word basis like WEAT.
By converting the original WEAT word lists into sentences with several contexts, the models can be better generalized and therefore can be tested for bias, as demonstrated by \citeauthor{may2019measuring}.
Similar to the WEAT lists of words, we translate and modify the sentences used in SEAT into Turkish.

\subsection{Turkish Bias Tests}


\begin{table*}[]
\centering
\footnotesize
    \begin{tabular}{llll} 

    \toprule
    \textbf{Type} &
    \textbf{Language} &
    \textbf{Target Concepts} & \textbf{Attributes}\\ 
    \midrule 

\multirow{8}{*}{
\rotatebox[origin=c]{90}{
\textbf{WEAT}
}
} 
& 
\multirow{4}{4em}{English in Original} & \textit{Male names}: & \textit{Career}: \\& & John, Paul, Mike,... & executive, management, professional,... \\ 
& &\textit{Female Names}: & \textit{Family}: \\ & & Amy, Joan, Lisa, Sarah,... & home, parents, children, marriage,... \\    

\rule{0pt}{3ex} 

&    \multirow{4}{4em}{Turkish Translation}& \textit{Male names}:  & \textit{Career}:\\ & & Mustafa, Orhan, Mehmet,... & yetkili, yönetim, profesyonel, şirket,... \\ 
& & \textit{Female Names}:  & \textit{Family}: \\  & & Zeynep, Elif, Selin, Fatma,... & ev, ebeveyn, çocuklar, aile,... \\

\midrule

\multirow{8}{*}{
\rotatebox[origin=c]{90}{\textbf{SEAT}}
} 
&
\multirow{4}{4em}{English in Original} & \textit{Male names}: `This is John.', & \textit{Career}: `This is an executive.', \\
& &  `Paul is here.', `Mike is a person.' &  `Management is here.' \\ 
& &\textit{Female Names}: `This is Amy.', & \textit{Family}: `This is a home.', \\
& &  `Lisa is here.', `Sarah is a person.'... &  `Family is here.' \\    

\rule{0pt}{3ex} 

&
    \multirow{4}{4em}{Turkish Translation}& \textit{Male names}: `Bu Mustafa.',  & \textit{Career}: `Bu bir yetkili.',\\
    & &  `Orhan burada.', `Mehmet orada.'... &  `Yönetim burada.', `Orada bir şirket var' \\ 
    &
& \textit{Female Names}: `Bu Zeynep.',  & \textit{Family}: `Bu bir ev.', \\ 
& & `Selin burada.', `Fatma orada.'... &  `Aile burada.', `Orada bir ebeveyn var.'\\

\midrule



\multirow{8}{*}{
\rotatebox[origin=c]{90}{\textbf{Double Binds (DB)}}
}
&

\multirow{4}{5em}{English in Original} & \textit{Male names}: & \textit{Competent and Achievement-oriented}:  \\
& &  `Paul is an engineer.',... &  `The engineer is productive.',... \\ 
& & \textit{Female Names}:& \textit{Incompetent and Not Achievement-oriented}: \\ & &  `Lisa is an engineer.',... &  `The engineer is unproductive.',... \\   

\rule{0pt}{3ex} 

 &
    \multirow{4}{5em}{Turkish Translation}&\textit{Male names}: & \textit{Competent and Achievement-oriented}:  \\
& &  `Mehmet bir mühendis..',... &  `Bu mühendis verimli.',... \\ 
& & \textit{Female Names}:& \textit{Incompetent and Not Achievement-oriented}: \\ 
& &  `Fatma bir mühendis.',... &  `Bu mühendis verimsiz.',... \\  

    \bottomrule
    \end{tabular}
\caption[]{Examples for Turkish data samples and their original English versions. The word-level and sentence-level tests (WEAT \& SEAT) compare the strength of the association
between the two target concepts and two attributes,
where all four are represented as sets of words or sentences.
Also, examples from the unbleached double bind test controlling for \textit{competence} and the Turkish translation are shown.}
\label{table:examplese}
\end{table*}


The English test data from \citet{may2019measuring} was used as a main reference to create our new Turkish datasets. The translation was manually performed by a co-author of this paper, who is a native Turkish speaker. 
The data provided by \citeauthor{may2019measuring}\footnote{\url{https://github.com/W4ngatang/sent-bias}} contains a total of 53 tests that include tests designed for evaluation of gender bias, social bias, racial bias and other biases.
The tests with person names were translated from US names into a diverse range of Turkish names that include both traditional and modern names. 
For each test with person names, an extra version using only religious names was created, which we hypothesized would result in a more significantly gender-biased result. 
In total, we created a total of 37 tests.\footnote{See the supplementary data for more details.}

\paragraph{Caliskan Tests.}
These tests examine whether the contextual word embedding methods reproduce the same biases that the GloVE word embedding models exhibited in \citet{caliskan2017semantics}. 
These biases correspond to past
social psychology studies of implicit associations
in human subjects \citep{greenwald1998measuring}.
The tests with numbers 1-3, 5, and 10 focus on language biases that are not relevant to this paper, such as the association between flowers and insects; hence, we discard them. 
\autoref{table:examplese} exemplifiese how the word-level and sentence-level Caliskan tests were translated into Turkish.

\begin{itemize}
    \item \textbf{Test 4: Relation between Turkish/Kurdish names and pleasant/unpleasant attributes}
    
    The original version developed for the US context examines the association of European and African American person names with pleasant/unpleasant attributes such as `love', `peace' and `hatred', `tragedy'. This stereotype is not a relevant one in the context of Türkiye and, therefore, in the context of Turkish language models. This test was adapted to the Turkish context by replacing European/African American person names with Turkish and Kurdish names. \citet{guvengez2020medyada} point out that in Turkish media there is a lot of discriminative and violent information against Kurdish people in the country. By replacing the person names in the target groups with Turkish and Kurdish names, our aim was to examine the ethnic bias in Turkish language models toward Kurdish people.
    \item \textbf{Test 6: Relation between male/female and career/family}
    \item \textbf{Test 8: Relation between male/female and science/art}
\end{itemize}
    
For the Caliskan Tests 6 and 8 the main version uses a person’s name. An alternative version uses general group terms (like `men', `boys', `girls' instead of `Michael', `Sarah' etc.) - this is annotated as version `b' in the rest of the paper and in supplementary data.
In the main version with person’s names, we used a diverse range of Turkish names (spanning from modern to old, religious, etc.). We created another version with only Muslim/traditional names, which we hypothesize will result in a higher gender-biased result. This is annotated as version `religious'. 
So, for each test, there is the main version, version `b', and version `religious'. Each of the Caliskan tests includes a word-level and a sentence-level test, which result in 6 tests for each. In total, 24 Calsikan tests were created.


\paragraph{Double Bind Tests.}
Women face many `double binds' (DB), that are contradictory or unsatisfiable expectations of femininity and masculinity \citep{stone2004fast}. 
If women clearly succeed in a job stereotypically associated with men, they are perceived as less likable and more hostile than
men in similar positions.
On the other hand, in the case of an ambiguous success scenario, they are perceived as incompetent and not achievement-oriented enough \citep{heilman2004penalties}.

\citet{may2019measuring} test this double bind in language models
by translating \citeauthor{heilman2004penalties}’s experiment to two
SEAT tests. 
In the first, they represent the two target concepts by names of women and men, respectively, in the single sentence template `\textit{<word> is
an engineer with superior technical skills.}'; the attributes are likable and non-hostile terms, based
on \citeauthor{heilman2004penalties}’s design, in the sentence template, `\textit{The engineer is <word>.}' Secondly, based on \citeauthor{heilman2004penalties}’s design, they use the abbreviated target concept sentence template `\textit{<word> is an engineer}' and fill
the previous attribute templates with terms representing `competence'. \citeauthor{may2019measuring} refer to these tests as semantically unbleached because it contains a context and the context contains important information about the bias. 
There are two variations of these tests: (1)  word-level tests in
which target concepts are names in isolation and
attributes are adjectives in isolation and (2) corresponding semantically bleached sentence-level tests.
Using these control conditions, it can be examined to which extent observed associations are attributable to gender independent of context. 
\autoref{table:examplese} shows how the unbleached sentence-level test was translated into Turkish.

\section{Results}

\begin{figure*}[htp]
    \centering
    \includegraphics[width=\linewidth]{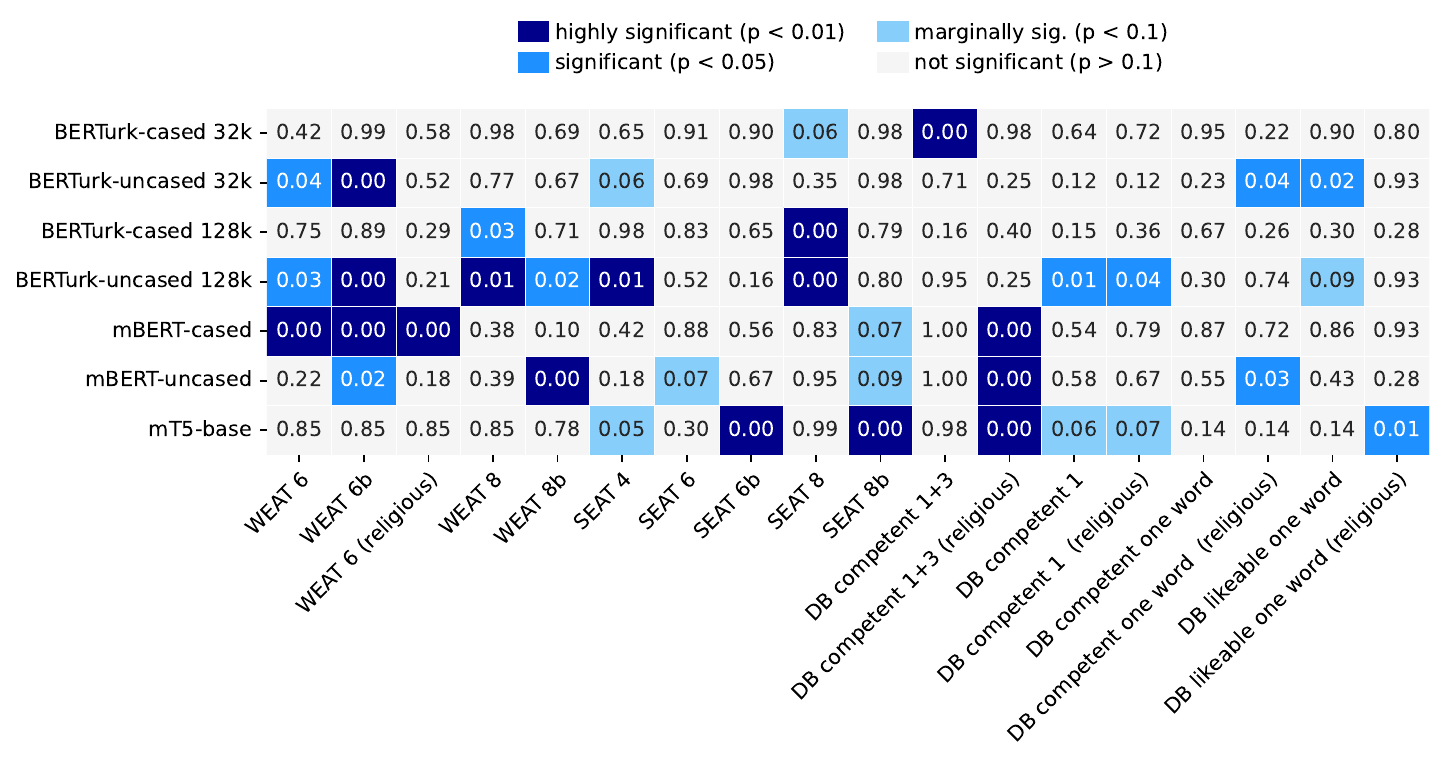}
    \caption{\label{fig:pvalues}
    P-values for a selected relevant subset of Turkish gender bias tests including WEAT, SEAT, and Double Bind (DB) tests. The tests using the group-terms are annotated with a `b' in the test name. WEAT tests illustrate that monolingual models elicit biased associations for given-name tests while the multilingual models demonstrate opposite behavior. Sentence-level SEAT4 test checking for the association of Turkish/Kurdish names with pleasant/unpleasent attributes, where high p-values indicate that the result is statistically insignificant.
    Double Bind tests show that the statistical significance increases when using traditional set of names instead of the mixed set of names.
    %
    }
    
\end{figure*}

\autoref{table:models} shows the list of the models we evaluated. 
Selecting these models, our aim was to compare the effect of multilingualism, the model size, and the pre-training corpus on the bias evaluation. 
Regarding the multilingual models, it is important to know the percentage of Turkish data in the whole pre-training corpus. 
Turkish counts as a low-resource language, and is relatively underrepresented in the training data. 
There is no public information on the exact percentage of Turkish datasets in the whole training data, we only know that English has approx. 33-times more representation.\footnote{\url{https://tensorflow.org/datasets/catalog/c4}}. 
Furthermore, the exact training data details of the mBERT models are not disclosed \footnote{\url{https://github.com/google-research/bert}}.

\begin{table}[]
\centering
\footnotesize
    \begin{tabular}{lll} 

    \toprule
     \textbf{Model (size, case)} & \textbf{Multilingual.}  & \textbf{Vocab.}\\ 
    \midrule 
 BERTurk (base, cased)  & mono    & 32k \\
 BERTurk (base, uncased)  & mono    & 32k  \\
  BERTurk (large, cased)  & mono     & 128k \\
  BERTurk (large, uncased)  & mono   & 128k\\
    mBERT (cased) & multi & 119k  \\
       mBERT (uncased) & multi & 119k \\
   mT5 (base) & multi & 250k  \\

    \bottomrule
    \end{tabular}
    \caption{List of the evaluated monolingual or multilingual language models with the vocabulary size.}
    \label{table:models}
\end{table}

We apply SEAT to seven LMs, as listed
in \autoref{table:models}, including four monolingual BERT \citeplanguageresource{stefan_schweter_2020_3770924} and two multilingual BERT models \citep{devlin2018bert}, and the base mT5 model \cite{raffel2020exploring, xue2020mt5}. For all models, we use pretrained checkpoints from Huggingface.

\subsection{Caliskan Tests}

\autoref{fig:pvalues} shows the statistical significance (p-value) for a subset\footnote{For all test results, see the GitHub repository.} of our word-level and sentence-level Caliskan tests (WEAT and SEAT). 
More precisely, we selected Test 6 associating
male/female names with career/family including the `religious' variation and Test 8 associating
male/female names with science/art attributes including the `group-term' variation, marked with \textit{b}.

Overall, as also pointed out in findings by \citet{may2019measuring, caliskan2017semantics}, sentence-level tests seem to elicit more biased associations than word-level tests.
For Caliskan Test 6 and 8,
sentence-level tests tend to elicit more significant
associations (p-value at 0.01) than word-level tests, while the latter
tend to have larger effect sizes. 
Especially for the Caliskan Test 8, the multilingual models showed much higher significance at the sentence-level, whereas no significant results were found at the word-level.
We discarded Caliskan Test 4, since the word-level test did not lead to any significant results for any of the models, whereas the sentence-level test picked up a more biased association for mT5 with a p-value at 0.05.

Furthermore, monolingual models seem to pick up no bias with gendered group terms (like `boys', `girls', etc.), but given-name tests have more biased associations. 
This outcome is in line with \citet{may2019measuring} who showed that names are more strongly associated with gender bias.
However, multilingual models demonstrated a reversed correlation: group-terms tests pick up more bias than person names, as illustrated in \autoref{fig:pvalues}.


\paragraph{Association of Kurdish/Turkish names with Pleasant/Unpleasant Attributes.}

The original test for English BERT by \citet{may2019measuring} that compared European and African American person names with the same attributes demonstrated a significant result with $p<10^{-4}$ and an effect size of 1.26 in the word-level test (WEAT4) and 0.69 in the sentence-level test (SEAT4).

Tests that compare Turkish names and Kurdish names for the association of pleasant/unpleasant terms (WEAT/SEAT 4) did not deliver significant results (p $\approx$ 1 in almost all tests) for any of the models, suggesting to accept the null hypothesis `No difference between Turkish names and Kurdish names in association to attributes Pleasant and Unpleasant'. The very low statistical significance suggests that there is almost zero correlation between those target concepts and attributes.

However, this result might be misleading. It is not clear whether the used Kurdish names occur in the corpus with sufficient frequency to pick up the bias existing the society. In further works, whether the used names have sufficient occurrence frequency in the corpus should be tested.

Among all the models tested, mT5 is the only model that picked up this bias with a p-value at 0.01 and an effect size of 0.35, while all the other models demonstrated a very low statistical significance  on this test. \autoref{fig:pvalues} illustrates the low p-value for mT5 (indicator for high statistical significance) and the contrasting very high p-values for other models. This could be explained by the different pretraining corpus of mT5.
mT5 is trained on mC4 corpus which has more political content, \citet{luccioni2021s} showed that mC4 has a lot of `toxic' content including a lot of hate speech and sexually explicit content.
On the other hand, mT5 is also trained on a Kurdish corpus. One could expect that the data from the Kurdish corpus should actually have had a counterbalancing effect on the ethnic bias deriving from the `toxic' embeddings in the Turkish corpus - as also hypothesizes by \citet{ahn2021mitigating}. However, the Turkish corpus in mC4 is 330x bigger than the Kurdish one (132,662,955 and 399,027 examples respectively)\footnote{\url{https://www.tensorflow.org/datasets/catalog/c4\#c4multilingual}, the specific units of these figures (TB or number of tokens etc.) are not given}. This could be the reason why the bias from the Turkish political data was still dominant and was not counterbalanced.

\paragraph{Association between male/female and career/family/science/art attributes.}

Overall, the monolingual models seem to have a `less biased' characteristic than the multilingual models.
However, \citet{ahn2021mitigating} suggest using multilingual BERT instead of monolingual BERT. 
The multiple languages used to train mBERT in one embedding space may have the effect of counterbalancing the ethnic bias in each monolingual BERT.

\citet{ahn2021mitigating} have shown that this phenomenon varies across the six languages they studied including English, Turkish, and Korean. 
This may have been due to the difference in the cultural context in the language corpus as language and culture
are entangled.
\citet{ahn2021mitigating} have shown that only for high-resource languages the multilingual model alone can mitigate the bias, or fine-tuning the multilingual model can effectively decrease the bias. 
However, they also propose other bias mitigation approaches that work for all languages, including low-resource ones,
such as the `contextual word alignment' approach \citep{ahn2021mitigating}, which is another bias mitigation method and is a better solution for low-resource languages.

Furthermore, \citet{wu2020all} present a statistical analysis to understand why mBERT does so poorly on some languages such as Korean and Turkish. They point to three factors that might affect the downstream task performance: pretraining Wikipedia size (WikiSize), task-specific supervision size, and
vocabulary size in task-specific data. 

\subsection{Double Binds}

\autoref{fig:pvalues} also shows the effect size and the statistical significance for a subset of our implementation of the Double Bind tests (DB) from \citet{may2019measuring}. 
We selected the \textit{competent} control tests, as the double bind tests controlling for `likable and not hostile' did not deliver any significant results. 
More precisely,  the word-level test, the bleached sentence-level test where the context is irrelevant to bias, and the unbleached sentence-level test where the context contains important information about the bias. For each of the tests, there is a \textit{religous} variation  that only uses traditional Turkish names instead of more modern ones.

Overall, for the models that demonstrated a statistically significant bias in these tests, the significance was increased when testing for religious names, meaning the double bind was more strongly picked up for traditional Turkish names, as depicted in \autoref{fig:pvalues}.


A so far unexplained observation is that there is a strong difference between the cased and uncased versions of the BERTurk models. The heatmap in \autoref{fig:pvalues} with all the test results visualizes this difference.
While the first and third rows (cased BERTurk models) did not demonstrate many significant results, the second and the fourth rows (uncased BERTurk models) show much higher statistically significant results. 
This suggests that uncased versions are more prone to bias than the cased versions. 
Since accent markers like  `ğ', `ü', `ş', `ö', `ç' are frequently used in the Turkish language, a reversed effect would have been more in line with our intuition: 
The uncased model removes all the accent markers \citep{devlin2018bert} and the uncased model would be expected to not capture the biased embeddings in the tests as much as the cased model. 
For future work focusing on this phenomenon, we suggest also making an uncased version of the dataset and running the models with these datasets for comparison.

While monolingual models and the multilingual BERT models seem to pick up the double bind, mT5 did not deliver many significant results for the double bind tests.

\paragraph{Competent and Achievement-oriented.}

\citet{may2019measuring} showed an evidence of the
double bind only in bleached, sentence-level \textit{competent}
control tests; suggesting women are associated
with incompetence independent of context. Our findings support this result: 
We found evidence of the double bind both in the bleached and the unbleached sentence-level tests. 
So, the context does not increase nor decrease the bias. 
Furthermore, with the growing size of the models, bias increases. 
The bleached sentence-level test has a marginal significance with \textit{p-value} at 0.1 and an effect size of 0.74 for the uncased BERTurk base model. 
For the larger model, the statistical significance increased to (\textit{p-value }< 0.01) and the effect size to 1.05. The effect size increased from 0.78 to 1.02 for the religous version of the same test. 
Moreover, the unbleached sentence-level test did not deliver any significant results for the smaller models and the larger BERTurk model had an effect size of 0.42 (significant at 0.05), and the religous version had an even bigger difference with an effect size of 0.50 (significant at 0.01), suggesting BERTurk picks up this bias at a higher significance level for tests with religious names.

\paragraph{Double Bind: Likeable and Not Hostile.}

When looking at the results of the multilingual models, an interesting observation is the effect of the selection of the used names. 
The double bind tests controlling for the `likability' with different verbosity (annotated as `1+3', and `1' in \autoref{fig:pvalues}) have alternating significant results for the cased and uncased version of mBERT.
The cased mBERT delivers highly significant results with p-value at 0.01 and effect sizes greater than  1.0 for the mixed group of names, while it does not pick up the same bias for the religious set of names. 
Interestingly, mBERT uncased demonstrates the opposite behavior. 

\section{Discussion}
The goal of this paper was to evaluate different common Turkish language models for their gender bias, and address the research gap for non-English languages in the content of LMs. 
We used existing bias evaluation frameworks on Turkish models by both translating existing English datasets and creating new ones designed to measuring gender bias in the context of Türkiye. 
We also extended the testing framework to evaluate Turkish models for their embedded ethnic bias toward Kurdish people.
Based on the test outcomes, we find possible relations of the picked-up biases to different model characteristics such as the model size, their multilingualism, and the training corpora.

Taking the testing framework of \citet{may2019measuring} as the main reference, we created a total of 37 tests that include tests designed for the evaluation of different gender stereotypes and ethnic bias toward Kurdish people. 
Our test results show that, overall, the monolingual models seem to have a `less biased' characteristic than the multilingual models. 
Although \citet{ahn2021mitigating} suggest using multilingual BERT instead of monolingual BERT to counterbalance the ethnic bias in monolingual BERT, mBERT has a reverse effect on low-resource languages like Turkish and demonstrates a poor performance on overall \citep{ahn2021mitigating,wu2020all}.

Furthermore, monolingual models seem to pick up no bias with gendered group terms (like `boys', `girls', etc.), but given-name tests have more biased associations. 
This outcome goes in line with \citet{may2019measuring} who showed that names are more strongly associated with gender bias.
However, multilingual models demonstrated a reversed correlation: group-terms tests pick up more bias than person names. Overall, as also pointed out in findings by \citet{may2019measuring, caliskan2017semantics}, sentence-level tests seem to elicit more biased associations than word-level tests.

The monolingual BERTurk models elicit stronger biased associations with the growing model size and vocabulary size. The different sizes of BERTurk have been elaborated on in \autoref{table:models}. 
This could be explained from two perspectives: 
The wording selection in the testing framework or the growing size of bias in larger language models.
Namely, this could be due to the selection of the words used in the tests, which do not have enough occurrence in the training corpus. 
During the translation and the creation of the Caliskan Tests, we did not focus whether the used names have sufficient occurrence frequency in the corpus. 
\citet{caliskan2017semantics, chavez-mulsa-spanakis-2020-evaluating} have pointed out that this is an important consideration when creating a new testing framework for LMs.
If the model did not come across the used vocabulary in its training, it self-evidently would not pick up any biased association. 
Another perspective is that larger models generally are more prone to bias existing in real-world data. \citet{tal2022fewer} point out that the bias increases with model size when measured using a prompt-based language task.
Generally, for the models that demonstrated a statistically significant bias, the significance was increased when testing for religious names in most of the tests -- confirming our hypothesis that LMs would pick up more strongly biased associations when using old and religion-rooted person names.

Lastly, we showed that the training corpus might have an impact on the bias the Turkish models pick up. 
The 'toxic' and political content in the mC4 corpus might be the reason why mT5 is the only model among all that has picked up the ethnic bias towards Kurdish people. 

\section{Conclusions and Future Work}

We have created a Turkish bias evaluation framework with diverse tests covering gender bias and ethnic bias  specifically designed for the Türkiye context, that is one of the first bias datasets in Turkish, if not the first. 
The 37 tests include in total translations of approximately 2,900 data entries and 19,300 words. 
Finally, as a contribution to the research community, the created datasets and code implementations were made publicly available to allow more researchers to work on this topic.\footnote{\githubUrl}

One of the limitations is that, although our methods were applied directly to Turkish, our test data were developed in English and adapted to Turkish. 
This means that our tests do not specifically take into account some linguistic features in Turkish that are not present in English. 
For instance, the use of additive suffixes and agglutination (a linguistic process in which words are formed by stringing together of morphemes: e.g. the word \textit{evlerinizden} -- `from your houses' in English -- consist of the morphemes \textit{ev-ler-iniz-den}), or the different forms of expression in Turkish dialects (like the dialects from the Blacksea or Egean regions) and regional languages such as the Laz language. 


Furthermore, some counterintuitive results cast doubt on the suitability of SEAT as a bias evaluation method. 
\citet{may2019measuring, kurita2019measuring} point out that cosine similarity might not be a suitable measure for contextual embeddings. 
In the future, the evaluation method should be analyzed more critically. 
Furthermore, SEAT's not-so-intuitive sensitivity to specific models and biases hints that the biases the SEAT tests uncover may not generalize beyond the specific words and phrases in our test data. 
This means that our results oppose the assumption that each set of words or phrases in our tests represents a consistent concept/attribute set (like Kurdish people or unpleasant attributes) for contextual word embeddings; therefore, we do not assume that the LMs will exhibit similar behavior for other potential elements of these concepts/attributes (other words or phrases representing, for instance, Kurdish people or unpleasant attributes). 
We highlight that our tests do not indicate the lack of bias, and merely examine whether bias exists in the specific test cases. 
There could be other non-tested cases where bias is present, thus further research into different bias evaluation techniques is recommended.


Finally, we recognize that our binary gender labels, deriving from the resources we use, do not reflect all gender identities. We hope that future works will extend our work to include non-binary gender identities in their research.

\subsubsection*{Acknowledgments}

The work presented in this paper has received funding from the German Federal Ministry for Economic Affairs and Climate Action (BMWK) through the project \href{https://opengpt-x.de/}{OpenGPT-X} (project no. 68GX21007D).

\bibliography{custom,languageresource}

\begin{thebibliography}{28}
\providecommand{\natexlab}[1]{#1}

\bibitem[{Ahn and Oh(2021)}]{ahn2021mitigating}
Jaimeen Ahn and Alice Oh. 2021.
\newblock Mitigating language-dependent ethnic bias in bert.
\newblock \emph{arXiv preprint arXiv:2109.05704}.

\bibitem[{Bender et~al.(2021)Bender, Gebru, McMillan-Major, and
  Shmitchell}]{bender2021dangers}
Emily~M Bender, Timnit Gebru, Angelina McMillan-Major, and Shmargaret
  Shmitchell. 2021.
\newblock On the dangers of stochastic parrots: Can language models be too big?
\newblock In \emph{Proceedings of the 2021 ACM Conference on Fairness,
  Accountability, and Transparency}, pages 610--623.

\bibitem[{Beukeboom and Burgers(2019)}]{beukeboom2019stereotypes}
Camiel~J Beukeboom and Christian Burgers. 2019.
\newblock How stereotypes are shared through language: a review and
  introduction of the aocial categories and stereotypes communication (scsc)
  framework.
\newblock \emph{Review of Communication Research}, 7:1--37.

\bibitem[{Blodgett et~al.(2020)Blodgett, Barocas, Daum{\'e}~III, and
  Wallach}]{blodgett2020language}
Su~Lin Blodgett, Solon Barocas, Hal Daum{\'e}~III, and Hanna Wallach. 2020.
\newblock Language (technology) is power: A critical survey of" bias" in nlp.
\newblock \emph{arXiv preprint arXiv:2005.14050}.

\bibitem[{Brown et~al.(2020)Brown, Mann, Ryder, Subbiah, Kaplan, Dhariwal,
  Neelakantan, Shyam, Sastry, Askell et~al.}]{brown2020language}
Tom Brown, Benjamin Mann, Nick Ryder, Melanie Subbiah, Jared~D Kaplan, Prafulla
  Dhariwal, Arvind Neelakantan, Pranav Shyam, Girish Sastry, Amanda Askell,
  et~al. 2020.
\newblock Language models are few-shot learners.
\newblock \emph{Advances in neural information processing systems},
  33:1877--1901.

\bibitem[{Buolamwini and Gebru(2018)}]{buolamwini2018gender}
Joy Buolamwini and Timnit Gebru. 2018.
\newblock Gender shades: Intersectional accuracy disparities in commercial
  gender classification.
\newblock In \emph{Conference on fairness, accountability and transparency},
  pages 77--91. PMLR.

\bibitem[{Caliskan et~al.(2017)Caliskan, Bryson, and
  Narayanan}]{caliskan2017semantics}
Aylin Caliskan, Joanna~J Bryson, and Arvind Narayanan. 2017.
\newblock Semantics derived automatically from language corpora contain
  human-like biases.
\newblock \emph{Science}, 356(6334):183--186.

\bibitem[{Ch{'a}vez~Mulsa and
  Spanakis(2020)}]{chavez-mulsa-spanakis-2020-evaluating}
Rodrigo~Alejandro Ch{'a}vez~Mulsa and Gerasimos Spanakis. 2020.
\newblock \href {https://www.aclweb.org/anthology/2020.gebnlp-1.6} {Evaluating
  bias in {D}utch word embeddings}.
\newblock In \emph{Proceedings of the Second Workshop on Gender Bias in Natural
  Language Processing}, pages 56--71, Barcelona, Spain (Online). Association
  for Computational Linguistics.

\bibitem[{Devlin et~al.(2018)Devlin, Chang, Lee, and
  Toutanova}]{devlin2018bert}
Jacob Devlin, Ming-Wei Chang, Kenton Lee, and Kristina Toutanova. 2018.
\newblock Bert: Pre-training of deep bidirectional transformers for language
  understanding.
\newblock \emph{arXiv preprint arXiv:1810.04805}.

\bibitem[{Gallegos et~al.(2023)Gallegos, Rossi, Barrow, Tanjim, Kim,
  Dernoncourt, Yu, Zhang, and Ahmed}]{gallegos2023bias}
Isabel~O. Gallegos, Ryan~A. Rossi, Joe Barrow, Md~Mehrab Tanjim, Sungchul Kim,
  Franck Dernoncourt, Tong Yu, Ruiyi Zhang, and Nesreen~K. Ahmed. 2023.
\newblock \href {https://arxiv.org/abs/2309.00770} {Bias and fairness in large
  language models: A survey}.
\newblock \emph{Preprint}, arXiv:2309.00770.

\bibitem[{Greenwald et~al.(1998)Greenwald, McGhee, and
  Schwartz}]{greenwald1998measuring}
Anthony~G Greenwald, Debbie~E McGhee, and Jordan~LK Schwartz. 1998.
\newblock Measuring individual differences in implicit cognition: the implicit
  association test.
\newblock \emph{Journal of personality and social psychology}, 74(6):1464.

\bibitem[{G{\"u}vengez et~al.(2020)G{\"u}vengez, Sa{\c{c}}, and
  Sert}]{guvengez2020medyada}
Serra G{\"u}vengez, Emircan Sa{\c{c}}, and G{\"u}lbeyaz Sert. 2020.
\newblock Medyada nefret s{\"o}ylemi ve ayr{\i}mc{\i} s{\"o}ylem 2019 raporu.

\bibitem[{Heilman et~al.(2004)Heilman, Wallen, Fuchs, and
  Tamkins}]{heilman2004penalties}
Madeline~E Heilman, Aaron~S Wallen, Daniella Fuchs, and Melinda~M Tamkins.
  2004.
\newblock Penalties for success: reactions to women who succeed at male
  gender-typed tasks.
\newblock \emph{Journal of applied psychology}, 89(3):416.

\bibitem[{Kraft(2021)}]{kraft2021triggering}
Angelie Kraft. 2021.
\newblock \emph{Triggering Models: Measuring and Mitigating Bias in German
  Language Generation}.
\newblock Ph.D. thesis, Master’s thesis, University of Hamburg.

\bibitem[{Kurita et~al.(2019)Kurita, Vyas, Pareek, Black, and
  Tsvetkov}]{kurita2019measuring}
Keita Kurita, Nidhi Vyas, Ayush Pareek, Alan~W Black, and Yulia Tsvetkov. 2019.
\newblock Measuring bias in contextualized word representations.
\newblock \emph{arXiv preprint arXiv:1906.07337}.

\bibitem[{Luccioni and Viviano(2021)}]{luccioni2021s}
Alexandra Luccioni and Joseph Viviano. 2021.
\newblock What’s in the box? an analysis of undesirable content in the common
  crawl corpus.
\newblock In \emph{Proceedings of the 59th Annual Meeting of the Association
  for Computational Linguistics and the 11th International Joint Conference on
  Natural Language Processing (Volume 2: Short Papers)}, pages 182--189.

\bibitem[{May et~al.(2019)May, Wang, Bordia, Bowman, and
  Rudinger}]{may2019measuring}
Chandler May, Alex Wang, Shikha Bordia, Samuel~R Bowman, and Rachel Rudinger.
  2019.
\newblock On measuring social biases in sentence encoders.
\newblock \emph{arXiv preprint arXiv:1903.10561}.

\bibitem[{Nadeem et~al.(2020)Nadeem, Bethke, and Reddy}]{nadeem2020stereoset}
Moin Nadeem, Anna Bethke, and Siva Reddy. 2020.
\newblock Stereoset: Measuring stereotypical bias in pretrained language
  models.
\newblock \emph{arXiv preprint arXiv:2004.09456}.

\bibitem[{{\"O}zaydinlik(2014)}]{ozaydinlik2014toplumsal}
Kevser {\"O}zaydinlik. 2014.
\newblock Toplumsal cinsiyet temelinde t{\"u}rkiye’de kad{\i}n ve
  e{\u{g}}itim.
\newblock \emph{Sosyal Politika {\c{C}}al{\i}{\c{s}}malar{\i} Dergisi}, (33).

\bibitem[{Parasurama and Sedoc(2021)}]{parasurama2021degendering}
Prasanna Parasurama and João Sedoc. 2021.
\newblock \href {https://doi.org/10.48550/ARXIV.2112.08910} {Degendering
  resumes for fair algorithmic resume screening}.
\newblock \emph{arXiv preprint}.

\bibitem[{Radford et~al.(2019)Radford, Wu, Child, Luan, Amodei, Sutskever
  et~al.}]{radford2019language}
Alec Radford, Jeffrey Wu, Rewon Child, David Luan, Dario Amodei, Ilya
  Sutskever, et~al. 2019.
\newblock Language models are unsupervised multitask learners.
\newblock \emph{OpenAI blog}, 1(8):9.

\bibitem[{Raffel et~al.(2020)Raffel, Shazeer, Roberts, Lee, Narang, Matena,
  Zhou, Li, Liu et~al.}]{raffel2020exploring}
Colin Raffel, Noam Shazeer, Adam Roberts, Katherine Lee, Sharan Narang, Michael
  Matena, Yanqi Zhou, Wei Li, Peter~J Liu, et~al. 2020.
\newblock Exploring the limits of transfer learning with a unified text-to-text
  transformer.
\newblock \emph{J. Mach. Learn. Res.}, 21(140):1--67.

\bibitem[{Schweter(2020)}]{stefan_schweter_2020_3770924}
Stefan Schweter. 2020.
\newblock \href {https://doi.org/10.5281/zenodo.3770924} {Berturk - bert models
  for turkish}.

\bibitem[{Stone and Lovejoy(2004)}]{stone2004fast}
Pamela Stone and Meg Lovejoy. 2004.
\newblock Fast-track women and the “choice” to stay home.
\newblock \emph{The Annals of the American Academy of Political and Social
  Science}, 596(1):62--83.

\bibitem[{Sun et~al.(2019)Sun, Gaut, Tang, Huang, ElSherief, Zhao, Mirza,
  Belding, Chang, and Wang}]{sun-etal-2019-mitigating}
Tony Sun, Andrew Gaut, Shirlyn Tang, Yuxin Huang, Mai ElSherief, Jieyu Zhao,
  Diba Mirza, Elizabeth Belding, Kai-Wei Chang, and William~Yang Wang. 2019.
\newblock \href {https://doi.org/10.18653/v1/P19-1159} {Mitigating gender bias
  in natural language processing: Literature review}.
\newblock In \emph{Proceedings of the 57th Annual Meeting of the Association
  for Computational Linguistics}, pages 1630--1640, Florence, Italy.
  Association for Computational Linguistics.

\bibitem[{Tal et~al.(2022)Tal, Magar, and Schwartz}]{tal2022fewer}
Yarden Tal, Inbal Magar, and Roy Schwartz. 2022.
\newblock Fewer errors, but more stereotypes? the effect of model size on
  gender bias.
\newblock \emph{arXiv preprint arXiv:2206.09860}.

\bibitem[{Wu and Dredze(2020)}]{wu2020all}
Shijie Wu and Mark Dredze. 2020.
\newblock Are all languages created equal in multilingual bert?
\newblock \emph{arXiv preprint arXiv:2005.09093}.

\bibitem[{Xue et~al.(2020)Xue, Constant, Roberts, Kale, Al-Rfou, Siddhant,
  Barua, and Raffel}]{xue2020mt5}
Linting Xue, Noah Constant, Adam Roberts, Mihir Kale, Rami Al-Rfou, Aditya
  Siddhant, Aditya Barua, and Colin Raffel. 2020.
\newblock mt5: A massively multilingual pre-trained text-to-text transformer.
\newblock \emph{arXiv preprint arXiv:2010.11934}.

\end{thebibliography}

\appendix

\end{document}